\documentstyle[12pt,mkstyle,twoside]{article}
\title{%
Identification and Interpretation of Belief Structure in
Dempster-Shafer Theory
}

\shorttitle{Identification of Belief Structure in DST}

\date{}
\newcommand{\Bem}[1]{}

\newcommand{\Prob}[2]{{ {\mbox{\gh Prob} ^{#2(#1)}} 
                         \atop {_{#1}} 
                     }}
\font\gh=eufm10 scaled \magstep1

\newcommand{\V}{{\bf V }}

\newcommand{\oantidot}{\overline{\odot}}

\begin{document}
\machetitel 
\begin{abstract}

Mathematical Theory of Evidence  
called also Dempster-Shafer Theory (DST) 
is known as a foundation
for reasoning when knowledge is expressed at various levels of
detail. Though much research effort has been committed to this
theory since its foundation, many questions remain open. One of
the most important open questions seems to be the relationship
between frequencies and the Mathematical Theory of Evidence.
The theory is blamed to leave frequencies outside (or aside of)
its framework. The seriousness of this accusation is obvious:
(1) no experiment may be run to compare the performance of
DST-based models of real world processes against real world
data, (2) data may not serve as foundation for construction of an
appropriate belief model.

In this paper we develop a frequentist interpretation  of the DST
bringing to fall the above argument against DST. 
An immediate consequence of it is the possibility to develop algorithms
acquiring automatically DST belief models from data. 
We propose three such algorithms for various classes of belief model
structures: for tree structured belief networks, for poly-tree belief
networks and for general type belief networks.
\end{abstract}

\section{Introduction}

The Dempster-Shafer Theory  or the Mathematical Theory of Evidence (DST) 
\cite{Shafer:76}, 
\cite{Dempster:67} 
 shows one of possible ways of application of mathematical probability for
subjective evaluation and is intended to be a generalization of bayesian
theory of subjective probability. \cite{Shafer:90ijar}. 

This theory offers a number of methodological advantages like: capability of
representing ignorance in a simple and direct way, compatibility with the
classical probability theory, compatibility of boolean logic and feasible
computational complexity  \cite{Ruspini:92ijar}. 

DST may be applied for (1) representation of incomplete knowledge, (2) belief
updating, (3) and for combination of evidence  \cite{Provan:92}.
DST covers the statistics of random sets and may be applied for representation
of incomplete statistical knowledge. Random set statistics is quite popular in
analysis of opinion polls whenever partial indecisiveness of respondents is
allowed \cite{Dubois:92}. 

Practical applications of DST include: integration of knowledge from
heterogeneous sources for object identification  \cite{deKorvin:93}, 
technical diagnosis under unreliable measuring devices  \cite{Durham:92}, 
medical applications: \cite{Gordon:90}, \cite{Zarley:88b}.
Relationships betwenn DST network reliability computation have been
investigated \cite{Provan:90}. %

In spite of indicated merits, DST experienced sharp criticism from many sides.
The basic line of criticism is connected with the relationship between the
belief function (the basic concept of DST) and frequencies. 

The problem of frequencies is not solely a scholar problem. It has significant
knowledge engineering (expressive power, sources of knowledge, 
 knowledge acquisition strategies, learning algorithms) and software
engineering
implications (internal representation, measure transformation procedures).
First of all one should realize that a computer-based advisory system is
rarely made for a single consultation. Hence we may (at least theoretically)
obtain a statistics of cases for which the system has been applied. Life
verifies also frequently enough the advices obtained from the advisory system.
Hence after a long enough time one may pose at least partially the question
whether or not the advices have been correct. A belief function (the basic
concept of DST) without a modest frequentist interpretation may be treated as
a void answer in this context.

Therefore numerous probabilistic interpretations have been attempted since the
early days of DST. 
Dempster \cite{Dempster:67} initiated interval interpretation of DST. \c{a} 
Kyburg \cite{Kyburg:87} showed  that the belief function may be
represented
by an envelop of a family of traditional probability  functions and claimed
that the behaviour of combining evidence via belief functions may be properly
explained in statistics under proper independence assumptions.
Hummel and  Landy \cite{Hummel:88} considered DST as a "statistics of expert
opinions" so that it 
 "contains nothing more than Bayes' formula applied
to Boolean assertions, .... (and) tracks multiple opinion as opposed to a
single probabilistic assessment".  

Pearl \cite{Pearl:90} and Provan \cite{Provan:90} considered belief functions
as "probabilities of provability". 

Still another view has been developed in connection with rough set theory
\cite{Grzymala:91}, 
 \cite{Skowron:93}, \cite{Skowron:93b}. Belief function is considered as the
lower approximation of the set of possible decisions.

Fagin and  Halpern \cite{Fagin:91} postulated probabilistic interpretation of
DST around lower and upper probability measures defined over a probability
structure (rather than space). 

Halpern and Fagin \cite{Halpern:92} proposed to treat Bel as a generalized
probability and proposed a rule of combination of evidence differing from the
one of Dempster and Shafer.  

The list of other attempts is quite long, still 
 a number of attempts to interpret belief functions in terms of probabilities
have failed so far to produce a fully compatible interpretation with DST - see
e.g. \cite{Kyburg:87}, \cite{Halpern:92}, \cite{Fagin:91} etc. Shafer
\cite{Shafer:90ijar} and Smets \cite{Smets:92}, in defense of DST, dismissed
every attempt to interpret DST frequentistically. Shafer stressed that
even modern (that meant bayesian) statistics is not frequentistic at all
(bayesian theory assigns subjective probabilities), hence frequencies be no
matter at all. Smets stated that domains of DST applications are those where
"we are ignorant of the existence of probabilities", 
and warns that DST is 
"not a 
model for poorly known probabilities" (\cite{Smets:92}, p.324). Smets states
further "Far too often, authors concentrate on the static component (how
beliefs are 
 allocated?) and discover many relations between TBM (transferable belief 
model of Smets) 
 and ULP (upper lower probability) models, inner and outer measures 
(Fagin and Halpern \cite{Fagin:89}), random sets (Nguyen \cite{Nguyen:78}), 
probabilities of provability 
 (Pearl \cite{Pearl:88}), probabilities of necessity (Ruspini 
\cite{Ruspini:86}) etc. But these authors 
usually do not explain or justify the dynamic component (how are beliefs 
updated?), that  is, how updating (conditioning) is to be handled (except in 
some cases by defining conditioning as a special case of combination). So I 
 (that is Smets) feel that these partial comparisons are incomplete, 
especially 
as all these interpretations lead to different updating rules." 
(\cite{Smets:92}, pp. 324-325).  

Shafer in  \cite{Shafer:90ijar} claims that probability theory developed over
last years from the old-style frequencies towards modern subjective
probability theory within the framework of bayesian theory. By analogy he
claims that the very attempt to consider relation between DST and frequencies
is old-fashioned and out of date and should be at least forbidden - for the
sake of progress of humanity. 
Wasserman opposes this view (\cite{Wasserman:92ijar}, p.371)
reminding  "major
 success story in Bayesian theory",  the exchangeability theory of 
de Finetti \cite{deFinetti:64}. It treats frequencies as special case of
bayesian belief.  "The Bayesian 
theory contains within it a definition of frequency probability and a 
description of the exact assumptions necessary to invoke that 
definition" \cite{Wasserman:92ijar}. Wasserman dismisses Shafer's suggestion
that probability relies on analogy of frequency. .

Though the need of a frequentist interpretation of DST is obvious, 
critical remarks of Smets cannot
be however ignored. Therefore, still another attempt of probabilistic
interpretation is made in this paper. Within this paper we assume strong
mutual relationship between the way an inference engine reasons, the way one
understands the inputs and outputs and the way knowledge is represented -m and
acquired. Section 2 reminds some basic terms of DST. Section 3 presents the
way we understand the belief function and the relationship between reasoning
system input and output. In section 4 we assume a particular type of inference
engine - Shenoy-Shafer method of local computations (for presentation of this
method the reader should consult the original paper of them
\cite{Shenoy:90}). Then we demonstrate how our understanding of belief
function and this particular inference mechanism imply knowledge
representation in terms of a new DS belief network. Section 5 and 6 show how
the results of section 4 lead to two learning algorithms recovering a
tree-structured and poly-tree-structured resp. belief network from data.
Section 7 makes use of independence results obtained in section 3 to develop
an algorithm recovering general type DS belief networks from data. Some
implications of this uniform view of belief functions, reasoning,
representation and acquisition are discussed in section 8. Conclusions are
summarized in section 9. 

\section{Basics of DST}

Let us first remind basic definitions of DST:
 
\begin{df} (see \cite{Provan:90})  
Let $\Xi$ be a finite  set of elements called elementary events. 
Any subset of $\Xi$ be a composite event. $\Xi$ be called also the 
frame of discernment.\\
A basic probability assignment function is any function m:$2^\Xi  \rightarrow
[0, 1]$ such that  $$  \sum_{A \in 2^\Xi } |m(A)|=1, \qquad
  m(\emptyset)=0, \qquad 
\forall_{A \in 2^\Xi} \quad  0 \leq  \sum_{A \subseteq B} m(B)$$
($|.|$ - absolute value.\\
      
A belief function be defined as Bel:$2^\Xi \rightarrow [0,1]$ so that 
 $Bel(A) = \sum_{B \subseteq A} m(B)$
A plausibility function be Pl:$2^\Xi \rightarrow [ 0,1]$  with 
$\forall_{A \in 2^\Xi} \  Pl(A) = 1-Bel(\Xi-A )$
A commonality function be Q:$2^\Xi-\{\emptyset\} \rightarrow [0,1]$ with 
 $\forall_{A \in 2^\Xi-\{\emptyset\}} \quad Q(A) = \sum_{A \subseteq B}
m(B)$ \end{df}

Furthermore, a Rule of Combination of two Independent Belief Functions 
$Bel_1$,
 $Bel_2$ Over the Same Frame of Discernment (the so-called Dempster-Rule),
denoted 
    $$Bel_{E_1,E_2}=Bel_{E_1} \oplus Bel_{E_2}$$ 
 is defined as follows: :
$$m_{E_1,E_2}(A)=c \cdot  \sum_{B,C; A= B \cap C} m_{E_1}(B) \cdot 
m_{E_2}(C)$$ (c - constant normalizing the sum of $|m|$ to 1)

Furthermore, let the frame of discernment $\Xi$ be structured in that it is
identical to cross product of domains $\Xi_1$, $\Xi_2$, \dots $\Xi_n$ of n
discrete variables $X_1, X_2, \dots X_n$, which span the space $\Xi$. Let
$(x_1, x_2, \dots x_n)$ be a vector in the space spanned by the variables 
$X_1,
 ,  X_2, \dots X_n$. Its projection onto the subspace spanned by variables 
$X_{j_1}, X_{j_2}, \dots X_{j_k}$ ($j_1, j_2,\dots j_k$ distinct indices from
the set 1,2,\dots,n) is then the vector $(x_{j_1}, x_{j_2}, \dots x_{j_k})$. 
$(x_1, x_2, \dots x_n)$ is also called an extension of $(x_{j_1}, x_{j_2},
\dots x_{j_k})$. A projection of a set $A$ of such vectors is the set
$A ^{\downarrow X_{j_1}, X_{j_2}, \dots X_{j_k}}$ 
 of
projections of all individual vectors from A onto $X_{j_1}, X_{j_2}, \dots
X_{j_k}$. A is also called an extension of $A ^{\downarrow X_{j_1}, X_{j_2},
\dots X_{j_k}}$. A is called the vacuous extension of $A ^{\downarrow
X_{j_1},
 X_{j_2}, \dots X_{j_k}}$  iff A contains all possible extensions of each
individual vector in $A ^{\downarrow X_{j_1}, X_{j_2}, \dots X_{j_k}}$ .
The fact, that A is a vacuous extension of B onto space $X_1,X_2,\dots\,
X_n$ is denoted by $A=B ^{\uparrow X_1,X_2,\dots\,X_n}$
\begin{df} (see \cite{Shenoy:90})
Let m be a basic probability assignment function on the space of discernment
spanned by variables   $X_1,X_2,\dots\,X_n$. $m ^{\downarrow X_{j_1},
X_{j_2}, \dots X_{j_k}}$ is  called  the  projection  of  m  onto 
subspace spanned by
$X_{j_1}, X_{j_2}, \dots X_{j_k}$ iff 
$$m ^{\downarrow X_{j_1}, X_{j_2}, \dots X_{j_k}}(B)= c \cdot
\sum_{A; B=A  ^{\downarrow X_{j_1}, X_{j_2}, \dots X_{j_k}} } m(A)$$
(c - normalizing factor)
\end{df}
\begin{df}  (see \cite{Shenoy:90})
Let m be a basic probability assignment function on the space of discernment
spanned by variables  $  X_{j_1},
X_{j_2}, \dots X_{j_k} $. $m ^{\uparrow X_1,X_2,\dots\,X_n}$ is called
the vacuous extension 
 of m onto superspace spanned by $X_1,X_2,\dots\,X_n$
iff 
$$m ^{\uparrow X_1, X_2, \dots X_n}(B ^{\uparrow X_1,X_2,\dots\,X_n})=m(B)$$

and $m ^{\uparrow X_1, X_2, \dots X_n}(A)=0$ for any other A. \\
We say that a belief function is vacuous iff $m(\Xi)=1$ and $m(A)=0$ for any A
different from $\Xi$.
\end{df}

Projections and vacuous extensions of Bel, Pl and Q functions are defined
with
respect to operations on m function. Notice that by convention if we want to
combine by Dempster rule two belief functions not sharing the frame of
discernment, we look for the closest common vacuous extension of their
frames of discernment without explicitly notifying it.

\begin{df} (See \cite{Shafer:90b}) Let B be a subset of $\Xi$, called 
evidence,
 $m_B$ be a basic probability assignment such that $m_B(B)=1$ and $m_B(A)=0$
for any A different from B. Then the conditional belief function $Bel(.||B)$
representing the belief function $Bel$ conditioned on evidence  B 
is defined
as: $Bel(.||B)=Bel \oplus Bel_B$. 
\end{df}

The subsequent definitions of hypergraphs and operations on them are due to
\cite{Shenoy:90}.

{\em Hypergraphs}: A nonempty set H of nonempty subsets of a finite set S be 
called 
a hypergraph on S. The elements of H be called hyperedges. Elements of S be 
called vertices. H and H' be both hypergraphs on S, then we call a 
hypergraph H' a {\em reduced hypergraph} of the 
hypergraph H, iff for every $h'\in H'$ also  $h'\in H$ holds, and for 
every  $h \in H$ there exists such a $h' \in H'$ that $h \subseteq h'$.
A hypergraph H {\em covers} a hypergraph H' iff for every $h'\in H'$ there 
exists such a $h\in H$ that $h'\subseteq h$.

{\em Hypertrees}: t and b be distinct hyperedges in a hypergraph H, $t \cap 
b\neq 
\emptyset$, and b contains every vertex of t that is contained in a hyperedge 
of H other than t; if $X\in t$ and $X\in h$, where $h\in H$ and $h\neq t$, 
then $X\in b$. Then we call t a twig of H, and we call b a branch for t. A 
twig may have more than one branch. 
We call a hypergraph a hypertree if there is an ordering of its hyperedges, 
say $h_1,h_2,...,h_n$ such that $h_k$ is a twig in the hypergraph 
$\{h_1,_h2,...,h_k\}$ whenever $2 \leq k \leq n$. We call  any  such 
ordering of 
hyperedges a hypertree construction sequence for the hypertree. The first 
hyperedge in the hypertree construction sequence be called the root of the 
hypertree construction sequence. 

{\em Variables and valuations}:Let {\V} be a finite set. The elements of {\V} 
are called 
variables. For each $h  \subseteq \V$ there is a set $VV_h$. The elements of  
 $VV_h$ are  called valuations. Let VV=$\bigcup \{ VV_h | h \subseteq \V \}$ 
be called the set of all valuations.

 In case of probabilities a valuation on h will be a non-negative, 
real-valued 
function on the set of all configurations  of h(a configuration on h is a 
vector of possible values of variables in h). In the belief function case  a 
valuation is a non-negative, real-valued function on the set of all     
subsets of 
configurations of h.

{\em Proper valuation}: for each $h \subseteq \V$ there is a subset $P_h$ of 
$VV_h$ 
elements of which are called proper valuations on h. Let P be the set of all 
proper valuations. 

{\em Combination}: We assume that there is a mapping $\odot: VV \times VV 
\rightarrow VV$ called combination such that:\\
(i) if G and H are valuations on g and h respectively, then $G \odot H$ is a 
valuation on $g \cup h$ \\
(ii) if either G or H is not a proper valuation then  $G \odot H$ is not a 
proper valuation \\
(iii) if both G and H are proper valuations then  $G \odot H$ may be or not 
be a proper valuation 

{\em Factorization}: Suppose A is a valuation on a finite set of variables \V,
and suppose HV is a hypergraph on \V. If A is equal to the combination of 
valuations of all hyperedges h  of HV then we say that A factorizes on HV.

\section{An Interpretation of DST}

Let
us assume that we know that objects of a  population  can 
be described by an  intrinsic attribute  X taking 
exclusively   one   of   the   n   discrete   values   from   its 
domain $\Xi=\{v_1,v_2,...,v_n\}$ . Let us  assume  furthermore 
that to obtain knowledge of the actual value taken by  an  object 
we must apply a measurement method (a system of tests) $M$  

\begin{df} \label{MDef}
$X$ be a set-valued attribute taking as its values non-empty
subsets of a finite domain $\Xi$.
By a measurement method 
of value of the attribute $X$
we understand a function:
 $$M: \Omega \times 2^\Xi \rightarrow \{TRUE,FALSE\}$$. 
where $\Omega$ is the set of objects, (or population of objects)
such that 
\begin{itemize}
\item
 $ \forall_{\omega; \omega \in \Omega} \quad
M(\omega,\Xi)=TRUE$ (X takes at least one of values from $\Xi$)
\item
 $ \forall_{\omega; \omega \in \Omega} \quad
M(\omega,\emptyset)=FALSE$ 
\item 
 whenever 
$M(\omega,A)=TRUE$
for $\omega \in \Omega$, $A \subseteq \Xi$
 then for any $B$ such that $A \subset B$ $M(\omega,B)=TRUE$   
holds,
 \item 
 whenever 
$M(\omega,A)=TRUE$
for $\omega \in \Omega$, $A \subseteq \Xi$ and if $card(A)>1$ then there 
exists  $B$, $B \subset A$ such that $M(\omega,B)=TRUE$ holds.
\item 
for every $\omega$ and every $A$
either  
$M(\omega,A)=TRUE$  or 
 $M(\omega,A)=FALSE$ (but never both).
 \end{itemize}
$M(\omega,A)$  tells us whether or not any of the elements of the set A 
belong to the actual value of the attribute $X$ for the object $\omega$.
 \end{df}

The measuring function M(O,A), if it takes the value TRUE,  
states for an object O and a set A of values from the domain of X  
that  the X 
takes for this object (at least) one of the  values  in A. 

 With each application of  the 
measurement  procedure  some  costs  be   connected,   increasing 
roughly with the decreasing size of the tested set A so that  we 
are ready to accept results of previous measurements in the  form 
of pre-labeling of the population. So 

\begin{df}
A {\em label} $L$ of an object $\omega \in \Omega$ is a subset of the domain
$\Xi$ of the attribute $X$. \\
A {\em labeling}  under the measurement method $M$  is a function $l: \Omega 
\rightarrow 2^\Xi$ such that for any object  $\omega \in \Omega$ either
$l(\omega)=\emptyset$ or $M(\omega,l(\omega))=TRUE$.\\
Each {\em labelled object}  (under the labeling $l$) 
consists of a 
pair $(O_j,L_j)$, $O_j$ - the j$^{th}$ object, $L_j=l(O_j)$ - its label.\\
By a {\em population  under the labeling $l$} we understand the predicate 
$P:\Omega \rightarrow \{TRUE,FALSE\}$ of the form 
$P(\omega)=TRUE \  iff \ l(\omega) \neq \emptyset$
(or alternatively, the set of objects  for which this predicate is true) \\
 If for every  object of the 
population the label is equal 
 to $\Xi$ then  we  talk  of  an  {\em unlabeled  population} (under the 
labeling $l$), otherwise of a {\em pre-labelled} one.
\end{df}

     Let  us  assume  that  in  practice  we  apply  a   modified 
measurement method 
$M_l$ being a function:

\begin{df} 
Let $l$ be a labeling under the measurement method $M$. 
Let us consider the population under this labeling.
The modified measurement method 
$$M_l:
 \Omega \times 2^\Xi \rightarrow 
\{TRUE,FALSE\}$$
where $\Omega$ is the set of objects, 
is is defined as  
$$M_l(\omega,A)= M(\omega,A \cap l(\omega) )$$  
(Notice that 
$M_l(\omega,A)=FALSE$ whenever $A \cap l(\omega)= \emptyset$.)
\end{df}

For a labeled object $(O_j,L_j)$  ($O_j$ - proper object, 
$L_j$  - 
its label)  and a set A of values from the domain of X, 
the modified measurement method tells us 
that $X$ takes one of the values in A if and only if it takes in fact 
a value from intersection of A and $L_j$.
 Expressed   differently,   we 
discard a priori any attribute not in the label.

Please pay attention also to the fact, that given a population P for which 
the measurement method $M$ is defined, the labeling $l$ (according to its 
definition) selects a subset of this population, possibly a proper subset, 
namely the population P'
under this labeling. 
$P'(\omega)=P(\omega) \land M(\omega,l(\omega))$. 
Hence also $M_l$ is defined possibly for the "smaller" 
population P' than $M$ is. \\

Let us    define  the  following  functions  referred  to  as 
labelled Belief, labelled Plausibility and labelled Mass 
Functions respectively for the labeled population P:
The predicate $\Prob{\omega}{P}\alpha(\omega)$ shall denote
the probability of truth of expression $\alpha(\omega)$ over $\omega$ given  
population $P(\omega)$.
 
\begin{df}
Let P be a population and $l$ its labeling. Then 

$$Bel_P    ^{M_l}(A)=\Prob{\omega}{P} \lnot M_l(\omega,\Xi-A)$$

$$Pl_P ^{M_l}(A)=\Prob{\omega}{P} M_l(\omega,A)$$

$$m_P ^{M_l}(A)=\Prob{\omega}{P} (\bigwedge_{B;B=\{v_i\}\subseteq A}
 M_l(\omega,B)
  \land \bigwedge_{B;B=\{v_i\}\subseteq \Xi-A} \lnot
 M_l(\omega,B))$$
\end{df}

\begin{th} 
$m_P ^{M_l}$, $Bel_P ^{M_l}$, $Pl_P ^{M_l}$ and $Q_P ^{M_l}$    are the mass,
belief, plausibility and commonality Functions in the sense of DST.
\end{th}

Let  us  now  assume  we  run  a  "(re-)labelling  process"   on   the 
(pre-labelled or unlabeled)
population P. 

\begin{df}
Let $M$ be a measurement method, $l$ be a labeling under this measurement
method, and P be a population under this labeling (Note that the population
may also be unlabeled).
The  {\em (simple) labelling  process}   on    
the
population P 
is defined as a functional 
$LP: 2^\Xi \times \Gamma \rightarrow \Gamma$, where $\Gamma$ is the set of  
all  possible labelings under $M$, 
such that for the given labeling $l$ and a given nonempty
set of attribute values $L$ ($L  \subseteq \Xi$), 
it delivers a new labeling $l'$ ($l'=LP(L,l)$) such that for every object
$\omega \in \Omega$: 

1. if  $M_l(\omega,L)=FALSE$ then  
$l'(\omega)=\emptyset$\\
(that is l' discards a
labeled 
 object $(\omega,l(\omega))$ if $M_l(\omega,L )=FALSE$ 

2. otherwise $l'(\omega)=l(\omega) \cap L $
(that is l' labels the object with $l(\omega) \cap L $ otherwise.
\end{df}

Remark: It is immediately obvious, that the population obtained as the sample 
fulfills the requirements of the definition of a labeled population.

The labeling process clearly induces from P another population P' (a 
population under the labeling $l'$) being a subset of P (hence perhaps 
"smaller" 
than P)   labelled  a 
bit differently. If we  retain  the  primary  measurement 
method M then a  new  modified  measurement  method 
$M_{l'}$ is induced by the new labeling. 

\begin{df} "labelling  process  function" 
$m ^{LP;L }: 2 ^\Xi \rightarrow [0,1]$:
 is defined as:
 $$m ^{LP;L }(L )=1$$  
$$\forall_{B;  B  \in  2^\Xi,B \ne L } m ^{LP;L }(B)=0$$
\end{df}

It is immediately obvious that:

\begin{th} 
 $m ^{LP;L }$ is a Mass Function in sense of DST.
\end{th}

Let  $Bel  ^{LP,L }$  be  the  belief  and  $Pl  ^{LP,L }$  be  the 
Plausibility corresponding to $m ^{LP,L }$. Now let  us  pose  the 
question: what is the relationship between $Bel_{P'} ^{M_{l'}}$, 
 $Bel_P ^{M_l}$,  and $Bel ^{LP,L }$. 

\begin{th} 
\label{thSimpleLab}
Let $M$ be a measurement function, $l$ a labeling, P a population under
this labeling. Let $L $ be a subset of $\Xi$. 
Let $LP$ be a labeling process and let $l'=LP(L ,l)$.
Let P' be a population under the labeling $l'$.
Then 
 $Bel_{P'} ^{M_{l'}}$ is a  combination  via  Dempster's  Combination 
rule of  $Bel ^{M_l}$,  and $Bel ^{LP;L }$., that is:
$$Bel_{P'} ^{M_{l'}} = Bel_P ^{M_l} \oplus Bel ^{LP;L }$$.
\end{th}

Let us define a more general (re-)labeling process. 
  Instead  of  a  single  set  of 
attribute  values  let  us  take  a  set  of  sets  of  attribute 
values $L ^1, L ^2, ...,L ^k$  (not  necessarily  disjoint)  and 
assign to each one a probability 
$m ^{LP, L ^1, L ^2, ...,L ^k}(A_i)$
of selection.

\begin{df}
Let $M$ be a measurement method, $l$ be a labeling under this measurement
method, and P be a population under this labeling (Note that the population
may also be unlabeled).
Let  us  take  a  set  of (not  necessarily  disjoint) nonempty sets  of  
attribute values $\{L ^1, L ^2, ...,L ^k\}$    and 
let us define the  probability of selection as a function
$m ^{LP, L ^1, L ^2, ...,L ^k}: 2 ^\Xi \rightarrow [0,1]$ such that
$$\sum_{A;A \subseteq \Xi}m ^{LP, L ^1, L ^2, ...,L ^k}(A)=1$$
$$\forall_{A; A \in \{ L ^1, L ^2, ...,L ^k\}} 
m ^{LP, L ^1, L ^2, ...,L ^k}(A)>0$$
$$\forall_{A; A \not\in \{ L ^1, L ^2, ...,L ^k\}} 
m ^{LP, L ^1, L ^2, ...,L ^k}(A)=0$$
 The  {\em (general) labelling  process}   on    
the
population P 
is defined as a (randomized) functional 
$LP: 2^{2^\Xi} \times \Delta
\times  \Gamma \rightarrow \Gamma$, where $\Gamma$ is the set 
of all  possible labelings under $M$, and $\Delta$ is 
a set of all possible probability of selection functions,
such that for the given labeling $l$ and a given 
 set  of (not  necessarily  disjoint) nonempty sets  of  
attribute values $\{L ^1, L ^2, ...,L ^k\}$    and 
a given probability of selection 
$m ^{LP, L ^1, L ^2, ...,L ^k}$
it delivers a new labeling $l"$ such that for every object
$\omega \in \Omega$:

1. a label L, element of the set $\{ L ^1, L ^2, ...,L ^k\}$ 
is sampled randomly according to the probability distribution 
$m ^{LP, L ^1, L ^2, ...,L ^k}$;
This sampling is done independently for each individual object,

2. if  $M_l(\omega,L)=FALSE$ then  
$l"(\omega)=\emptyset$\\
(that is l" discards an object $(\omega,l(\omega))$ if 
$M_l(\omega,L )=FALSE$ 

3. otherwise $l"(\omega)=l(\omega) \cap L $
(that is l" labels the object with $l(\omega) \cap L $ otherwise.)
\end{df}

Again we obtain another ("smaller") population P" under the labeling $l"$  
labelled 
 a bit differently. Also a  new  modified  measurement  method 
$M_{l"}$ is induced by the "re-labelled" population.  
Please notice, that $l"$ is not derived deterministicly. 
Another run of the general (re-)labeling process LP may result in a different
final labeling of the population and hence a different subpopulation under 
this new labeling.

Clearly:

\begin{th} 
$m ^{LP,L ^1,...,L ^k}$ is a Mass Function in sense of DST.
\end{th}

Let   $Bel   ^{LP;L ^1,...,L ^k}$   be   the   belief    and    $Pl 
^{LP,L ^1,...,L ^k}$  be  the 
Plausibility corresponding to $m ^{LP,L ^1,...,L ^k}$. Now let  us  pose  the 
question: what is the relationship between $Bel_{P"} ^{M_{l"}}$, 
 $Bel_P ^{M_l}$,  and $Bel ^{LP,L ^1,...,L ^k}$. 

\begin{th} 
Let $M$ be a measurement function, $l$ a labeling, P a population under
this labeling. 
Let $LP$ be a generalized labeling process and let $l"$
be the result of application of the $LP$ for the set
of labels from the set $\{ L ^1, L ^2, ...,L ^k\}$ 
 sampled randomly according to the probability distribution 
$m ^{LP, L ^1, L ^2, ...,L ^k}$;.
Let P" be a population under the labeling $l"$.
Then 
The expected value 
over the set of all possible resultant labelings $l"$ (and hence
populations P") 
(or, more precisely, value vector) of 
$Bel_{P"} ^{M_{l"}}$ is a  combination  via  Dempster's  Combination 
rule of  $Bel_P ^{M_l}$,  and $Bel ^{LP,L ^1,...,L ^k}$., that is:
$$E(Bel_{P"} ^{M_l'}) = Bel_P ^{M_l} \oplus Bel ^{LP,L ^1,...,L ^k}$$.
\end{th}

Let us now introduce the notion of quantitative independence for DS-Theory. 
We will fix the measurement method M we use and the population P we consider
so that respective indices will be usually dropped.

\begin{df}
Two variables $X_1,X_2$  are (mutually, marginally) independent when for 
objects of the population
knowledge of the truth value of $M_l ^{\downarrow X_1}
(\omega,A ^{\downarrow X_1})$ for all 
$A \subseteq \Xi_1 \times \Xi_2$ does not change our prediction capability
of the values  of $M_l ^{\downarrow X_2}
(\omega,B ^{\downarrow X_2})$ for any
$B \subseteq \Xi_1 \times \Xi_2$, that is
$$\Prob{\omega}{P}( M_l ^{\downarrow X_2}(\omega,B ^{\downarrow X_2})=
\Prob{\omega}{P}( M_l ^{\downarrow X_2}(\omega,B ^{\downarrow X_2})
|   M_l ^{\downarrow X_1}(\omega,A ^{\downarrow X_1})  )
 $$
 \end{df}

\begin{th}
If variables $X_1,X_2$ are quantitatively independent, then 
for any $B \subseteq \Xi_2$, $A \subseteq \Xi_1$
$$
 m  ^{\downarrow X_2}(B) \cdot  m ^{\downarrow X_1}(A) =
\sum_{F; F ^{\downarrow X_1}=A,F ^{\downarrow X_2}=B} m(F) $$
\end{th}

\begin{th}
If variables $X_1,X_2$ are quantitatively independent, then 
for any $B \subseteq \Xi_2$, $A \subseteq \Xi_1$
$$
 Bel  ^{\downarrow X_2}(B) \cdot  Bel ^{\downarrow X_1}(A) = Bel(A \times B)$$
\end{th}

This actually expresses the relationship between marginals of two 
independent variables and their joint belief  distribution.  This 
relationship has one dismaying  aspect:  in  general,  we  cannot 
calculate  the  joint  distribution  from  independent  marginals 
(contrary to our  intuition  connected  with  joint  probability 
distribution). 

In practical settings, however, we frequently have to do
 with some kind 
of composite measurement method, that is:

\begin{df} Two variables $X_1,X_2$ are measured compositely iff 
for $A \subseteq \Xi_1, B \subseteq \Xi_2$:
$$M(\omega,A \times C) = M(\omega, A \times \Xi_2) \land
M(\omega, \Xi_1 \times C) $$
and whenever $M(\omega, B)$ is sought,
$$M(\omega, B) = \bigvee_{A,C; A \subseteq \Xi_1, C \subseteq \Xi_2,
A\times C \subseteq B}
 M(\omega,A \times B)$$
\end{df}

Under these circumstances, it is easily shown that 
 whenever $m(B) > 0 $, then there  exist 
A and C such that: $B = A \times C$.

So we obtain:

\begin{th}
If variables $X_1,X_2$ are quantitatively independent and measured
compositely, then 
$$m(A \times C)= m ^{\downarrow X_1}(A) \cdot   m ^{\downarrow X_2}(C) 
$$
\end{th}

Hence the Belief function can be calculated from Belief functions 
of independent variables under composite measurement.:

\begin{th}
If variables $X_1,X_2$ are quantitatively independent and measured
compositely, then 
$$Bel=Bel ^{\downarrow X_1} \oplus  Bel ^{\downarrow X_2}
$$
\end{th}

Let us justify now the notion of empty extension:

\begin{df}
The joint distribution over $X=X_1 \times X_2$
in variables $X_1,X_2$  is 
independent of the variable $X_1$  when for objects of the population
for every A,$A \subseteq \Xi_1 \times \Xi_2$
knowledge of the truth value of $M_l ^{\downarrow X_1}
(\omega,A ^{\downarrow X_1})$ 
 does not change our prediction capability
of the values  of $M_l(\omega,A )$, that is
$$\Prob{\omega}{P} (M_l (\omega,A ))=
\Prob{\omega}{P}( M_l (\omega,A) 
|   M_l ^{\downarrow X_1}(\omega,A ^{\downarrow X_1}) )
 $$
 \end{df}
\begin{th}
The joint distribution over $X=X_1 \times X_2$
in variables   $X_1,X_2$,  measured
compositely,
 is 
independent of the variable $X_1$  only if 
$m ^{\downarrow X_2}(\Xi_2)=1$
that is the whole mass of the marginalized distribution onto $X_2$ is 
concentrated at the only focal point 
$\Xi_2$. 
\end{th}

\begin{th}
If for $X=X_1 \times X_2$ 
$Bel=(Bel ^{\downarrow X_2})^{\uparrow X}$
that is 
Bel is the empty extension of some Bel defined  only over $X_2$,
then the Bel is independent of the variable $X_2$.\\
If for a Bel over  $X=X_1 \times  X_2$  with  $X_1,X_2$  measured 
compositely
Bel is independent of $X_2$, then $Bel=(Bel ^{\downarrow X_2})^{\uparrow X}$.
\end{th}

\begin{df}
Let Bel be defined over $X_1 \times X_2$.
We shall speak that Bel is {\em compressibly independent} of $X_2$ iff 
$Bel=(Bel ^{\downarrow X_1}) ^{\uparrow X_2}$. 
\end{df}

REMARK: $m  ^{\downarrow X_1}(\Xi_1)=1$ does not imply empty extension as 
such, especially for non-singleton values of the variable $X_2$. As previously
with marginal independence, it is the composite measurement that 
makes the empty extension a practical notion.\\

Let us introduce a concept of conditionality related  to  the 
above definition of independence.  Traditionally,  conditionality 
is introduced to obtain a kind of independence between  variables 
de facto on one another. So let us define that:

\begin{df}
For discourse spaces of the form $X=X_1 \times ... \times X_n$
we define (anti-)conditional belief function 
$Bel ^{X | X_i}(A)$ as 
 $$Bel=Bel ^{\downarrow X_i} \oplus  Bel ^{X | X_i}$$
\end{df}

Let us notice at this point that the (anti-)conditional belief as defined
above does not need to be unique, hence we have here a kind of pseudoinversion
of the $\oplus$ operator. Furthermore, the conditional belief does not  need
to 
be a belief function at all, because some focal points m may be negative. 
But it is then the pseudo-belief function in the sense of the DS-theory as 
the Q-measure remains positive. 
Please recall the fact that if $Bel_{12}=Bel_1 \oplus Bel_2$ then
$Q_{12}(A)=c\cdot Q_1(A)\cdot Q_2(A)$, c being a proportionality factor
 (as all supersets of a set are contained in all intersections of its
supersets 
and vice versa). Hence also for 
our conditional belief definition: 
 $$Q(A)=c \cdot  (Q ^{\downarrow X_i})  ^{\uparrow X}(A) \cdot  Q ^{X |
X_i}(A)$$ We shall talk later of unnormalized conditional belief iff\\
 $$ Q_*      ^{X | X_i}(A) = Q(A)/(Q ^{\downarrow X_i})  ^{\uparrow X}(A) $$ 
 Let us now reconsider the problem of independence, this time of a 
conditional distribution of $(X_1 \times X_2 \times X_3 | X_1 \times X_3)$ 
from the third variable $X_3$.

\begin{th}
Let $X =X_1 \times X_2 \times X_3 $ and let $Bel$ be defined 
 over X.
Furthermore let $Bel ^{X|X_1 \times X_3}$ be a conditional Belief conditioned 
on variables $X_1,X_3$. Let this conditional distribution be
 compressibly 
 independent of 
$X_3$. Let $Bel ^{\downarrow X_1 \times X_2}$ be the projection of $Bel$ onto 
the subspace spanned by $X_1,X_2$. Then there exists 
 $Bel ^{\downarrow X_1 \times X_2 | X_1}$  being a conditional belief of that 
projected belief conditioned on the variable $X_1$ such that this  $Bel 
^{X|X_1 \times 
X_3}$  is the empty extension of  $Bel ^{\downarrow X_1 \times X_2 | X_1}$ 
 $$Bel ^{X|X_1 \times X_3} =
  (Bel ^{\downarrow X_1 \times X_2 | X_1}) ^{\uparrow X}$$
\end{th}

Let us notice that under the conditions of the above theorem

 $$ Bel = 
Bel ^{X|X_1 \times X_3}\oplus Bel ^{\downarrow X_1 \times X_3 } =
  Bel ^{\downarrow X_1 \times X_2 | X_1} 
\oplus Bel ^{\downarrow X_1 \times X_3 }
$$

and hence for any $Bel ^{\downarrow X_1 \times X_3 | X_1}$
$$Bel =  
  Bel ^{\downarrow X_1 \times X_2 | X_1}
\oplus Bel ^{\downarrow X_1}
\oplus Bel ^{\downarrow X_1 \times X_3 | X_1 }
$$

and therefore
$$Bel =  
  Bel ^{\downarrow X_1 \times X_2} 
\oplus Bel ^{\downarrow X_1 \times X_3 | X_1 }
$$
This means that whenever the conditional 
$Bel ^{X_1 \times X_2 \times X_3|X_1 \times X_3}$ 
is compressibly independent of $X_3$, 
then there exists a 
conditional 
$Bel ^{X_1 \times X_2 \times X_3|X_1 \times X_2}$ 
compressibly independent of $X_2$.
But this fact  combined with the previous theorem results in:

\begin{th}
Let $X =X_1 \times X_2 \times X_3$ and let $Bel$ be defined 
over X.
Furthermore let $Bel ^{X|X_1 \times X_3}$ be a conditional Belief conditioned 
on variables $X_1,X_3$. Let this conditional distribution be 
compressibly independent of 
$X_3$. 
Then the empty extension onto $X$ of any 
 $Bel ^{\downarrow X_1 \times X_2 | X_1}$  being a conditional belief of 
projected belief conditioned on the variable $X_1$ 
is a conditional belief function of $X$  conditioned 
on variables $X_1,X_3$. Hence for every $A\subseteq \Xi$
$$ \frac{ Q(A) }
{ Q ^{\downarrow X_1 \times X_3}(A ^{\downarrow X_1 \times X_3} ) }
=  \frac { Q ^{\downarrow X_1 \times X_2}(A ^{\downarrow X_1 \times X_2} ) }
{ Q ^{\downarrow X_1}(A ^{\downarrow X_1} ) }
$$
\end{th}


In this way we obtained some sense of conditionality suitable for 
decomposition of a joint belief distribution. 

\subsection{Independence from Data}

The preceding sections defined precisely what is meant by marginal 
independence of two variables in terms of the relationship between marginals 
and the joint distribution, as well as concerning the independence of a joint 
distribution from a single variable.\\

For the former case we can establish frequency tables  with  rows 
and columns 
corresponding to cardinalities of focal points of the first and the second 
marginal, and inner elements being cardinalities from the respective sum on 
DS-masses of 
the joint distribution. Clearly, cases falling into different inner 
categories of the table are different and hence $\chi ^2$ test is 
applicable.\\
The match can be $\chi ^2$-tested. The following 
formula should be followed for calculation 
$$\sum_{A;A \subseteq \Xi_1, m ^{\downarrow X_1}(A)>0}
  \sum_{B;B \subseteq \Xi_2, m ^{\downarrow X_2}(B)>0}
\frac { ((
\sum_{C;C \subseteq \Xi, A = C ^{\downarrow X_1}
B=C  ^{\downarrow X_2}}
m(C))
- m ^{\downarrow X_1}(A)\cdot  m ^{\downarrow X_2}(B))^2
}
{ m ^{\downarrow X_1}(A)\cdot  m ^{\downarrow X_2}(B)}$$

The number of df is calculated as
$$ (card(\{A;A \subseteq \Xi_1, m ^{\downarrow X_1}(A)>0\}) -1)\cdot 
(card(\{B;B \subseteq \Xi_2, m ^{\downarrow X_2}(B)>0\})-1)$$


In case of independence of a distribution from one variable one needs to 
calculate the marginal of the distribution of that variable, say $X_i$.
Then the measure of discrepancy from the assumption of independence is given 
as:
$$ 1 - m ^{\downarrow X_i}(\Xi_i)$$
Statistically we can test, based on Bernoullie distribution, what is the 
lowest possible and the highest possible value of $ 1 - m ^{\downarrow 
X_i}(\Xi_i)$ for a given significance level of the true underlying 
distribution.\\

 \subsection{Conditional Independence from Data}

In  case of 
independence between the conditional distribution and one of conditioning
variables, however, it is useless to calculate the pseudoinversion of 
$\oplus$, as we are working then with a population and a sample the size of 
which is not properly defined (by the "anti-labeling").
  But we can build the contingency table of the unconditional joint 
distribution for the independent variable on 
the one hand  and the remaining variables on the other hand, and compare the
respective cells on how do they match the distribution we would obtain 
assuming the  independence. The number of degrees of freedom for the      
$\chi ^2$ 
test would then be the number of focal points of the joint distribution,
minus 
the number of focal points within each of the multi-variable marginals plus 
one (for covering twice the total sum of 1 on all focal points).

 If
we test conditional independence of variables $X$ and $Y$ on the set of
variables $Z$, then we have to compare empirical distribution $Bel
^{\downarrow X,Y,Z}$ with $Bel ^{\downarrow X,Z|Z} \oplus Bel ^{\downarrow Y,
Z|Z} \oplus Bel ^{\downarrow Z}$. The traditional $\chi ^2$ statistics is
computed (treating  the latter distribution as expected one).  If the
hypothesis of equality is rejected on significance level $\alpha=0.05$ then
X and Y are considered dependent, otherwise independent. 

\section{A Concept of Belief Network} 
The axiomatization system of Shenoy/Shafer refers to the notion of
factorization along a hypergraph.
On the other hand other authors insisted on a decomposition 
into a belief network. We investigate below implications
of this disagreement. BEl shall denote the general belief function
as considered in \cite{Shenoy:90}, the DS belief function Bel
and the probability are specialization of.
%
%
\begin{df} \cite{Klopotek:93f}
We define               a mapping $\oantidot: VV \times VV 
\rightarrow VV$ called decombination such that: 
if $BEL_{12}=BEL_1 \oantidot BEL_2$ then $BEL_1=BEL_2 \odot BEL_{12}$.
 \end{df}
In case of probabilities, decombination means memberwise division: 
$Pr_{12}(A)=Pr_1(A)/Pr_2(A)$. In case of DS pseudo-belief functions it means 
the operator $\ominus$ yielding a DS pseudo-belief function such that: 
whenever $Bel_{12}=Bel_1 \ominus Bel_2$ 
then $Q_{12}(A)=c \cdot Q_1/Q_2$. Both for probabilities and for DS belief 
functions decombination may be not uniquely determined. Moreover, for DS 
belief functions not always a decombined DS belief function will exist. But   
          the domain of DS pseudo-belief functions       is closed under this 
operator. We claim here without a proof (which is simple) that DS 
pseudo-belief 
functions fit the axiomatic framework of Shenoy/Shafer. Moreover, we claim 
that if an (ordinary) DS  belief  function  is  represented  by  a 
factorization in 
DS pseudo-belief functions, then any propagation of uncertainty yields the 
very 
same results as when it would have been factorized into ordinary DS belief 
functions. 
\begin{df}  \cite{Klopotek:93f}
By anti-conditioning $|$ of a belief function $BEL$ on a set of variables
$h$ 
we understand the transformation: $BEL ^{|h}= BEL \oantidot BEL ^{\downarrow 
h}$. 
\end{df}
Notably, anti-conditioning means in case of probability functions proper 
conditioning. In case of DS pseudo-belief functions the operator $|$
means the  DS anti-conditioning from previous section.
It has 
meaning entirely different from traditionally used 
Shafer's 
notion of conditionality 
(compare 
\cite{Klopotek:93p4}) - anti-conditioning is a technical term used
exclusively 
for valuation of nodes in belief networks. Notice: some other authors 
e.g. \cite{Cano:93} recognized also the necessity of introduction of two 
different notions in the context of the Shenoy/Shafer axiomatic framework 
(compare a priori and a posteriori conditionals in \cite{Cano:93}). 
\cite{Cano:93} introduces 3 additional axioms governing the 'a priori' 
conditionality to enable propagation with them.  
Our 
anti-conditionality is bound only to the assumption of executability of the 
$\oantidot$ operation and does not assume any further properties of it. 
Let 
us define now the general notion of belief networks
%
%
\begin{df}  \cite{Klopotek:93f}
 A 
belief 
 network is a pair (D,BEL) where D is a dag (directed acyclic graph)
and BEL  is a belief 
distribution called the {\em underlying distribution}. Each node i in D 
corresponds to a variable $X_i$  in BEL, a set of nodes I corresponds to a 
set of variables $X_I$ and $x_i, x_I$
 denote values drawn from the domain of $X_i$ 
 and from the (cross product) domain of $X_I$ respectively. Each node in the 
network  is regarded as a storage cell for any  distribution 
$BEL ^{\downarrow \{X_i\} \cup X_{\pi (i)} |  X_{\pi (i)} }$
 where $X_{\pi (i)}$ is a set of nodes corresponding to 
the 
parent nodes $\pi(i)$ of i.  The underlying distribution represented by a 
 belief network is computed via:
$$BEL  = \bigodot_{i=1}^{n}BEL ^{\downarrow \{X_i\} \cup X_{\pi (i)} |  
X_{\pi (i)} } $$
\end{df}
Please notice the local character of valuation of a node:
to valuate the node $i$ corresponding to variable $X_i$ only 
the marginal $BEL ^{\downarrow \{X_i\} \cup X_{\pi (i)}}$ needs to be known 
(e.g. from data) and not the entire belief distribution.

There exists a straight forward transformation of a belief network structure
into a hypergraph, and hence of 
a belief network into a hypergraph:
for every node i of the underlying dag define a hyperedge as the set
$\{X_i\} \cup X_{\pi(i)}$; then the valuation of this hyperedge define as
$BEL ^{\downarrow \{X_i\} \cup X_{\pi(i)} | X_{\pi(i)}}$. We say that the 
hypergraph obtained in this way is {\em induced} by the belief network.
%
%

Let us consider now the inverse operation: transformation of a valuated
hypergraph into a belief network.
As the first  stage we consider structures of a hypergraph and of a
belief network (the underlying dag). we say that a belief network is 
{\em compatible} with a hypergraph 
iff the reduced set of hyperedges induced by    the belief network is 
identical with the reduced hypergraph. 

\begin{Bsp} 
Let us consider the following hypergraph 
\{\{A,B,C\}, \{C,D\}, \{D,E\}, \{A, E\}\}.
the following belief network structures are compatible with this hypergraph:
\{$A,C\rightarrow B$, $C\rightarrow D$, $D\rightarrow E$, $E\rightarrow A$\}
\{$A,C\rightarrow B$, $D\rightarrow C$, $D\rightarrow E$, $E\rightarrow A$\},
\{$A,C\rightarrow B$, $D\rightarrow C$, $E\rightarrow D$, $E\rightarrow A$\},
\{$A,C\rightarrow B$, $D\rightarrow C$, $E\rightarrow D$, $A\rightarrow E$\}.
\end{Bsp}


\begin{Bsp}
Let us consider the following hypergraph 
\{\{A,B,C\}, \{C,D\}, \{D,E\}, \{A, E\}, \{B,F\}, \{F,D\}\}.
No belief network structure is compatible with it.
\end{Bsp}
The missing compatibility is connected with the fact that
a hypergraph may represent a cyclic graph. 
Even if a compatible belief network has been found we may have troubles with 
 valuations. In Example 1 an unfriendly valuation of hyperedge 
\{A,C,B\} may require an edge AC in a belief network representing the same 
distribution, but  it will  make 
the 
hypergraph incompatible (as e.g. hyperedge \{A,C,E\} would be induced). This 
may be demonstrated as follows:
\begin{df}   \cite{Klopotek:93f}
If $X_J,X_K,X_L$ are three disjoint sets of variables of a distribution BEL, 
then $X_J,X_K$ are said to be conditionally independent given $X_L$ (denoted 
\linebreak 
$I(X_J,X_K |X_L)_{BEL}$ iff 
 $$BEL ^{\downarrow X_J \cup X_K \cup X_L |  X_L} 
 \odot  BEL ^{\downarrow   X_L } =
 BEL ^{\downarrow X_J  \cup X_L |  X_L} \odot 
 BEL ^{\downarrow X_K \cup X_L |  X_L} 
 \odot  BEL ^{\downarrow   X_L } $$
%
$I(X_J,X_K |X_L)_{BEL}$ is called a {\em 
conditional independence statement}
\end{df}
Let $I(J,K|L)_D$ denote d-separation in a graph \cite{Geiger:90}.:
\begin{th} \label{IDIBEL} 
 \cite{Klopotek:93f}
Let $BEL_D=\{BEL|$(D,BEL) be a belief network\}. Then:\\
$I(J,K|L)_D$ iff $I(X_J,X_K |X_L)_{BEL}$ for all $BEL \in BEL_D$.
\end{th}
Now we see in the above example that nodes D and E d-separate nodes A and C. 
 Hence within any belief network based on one of the three dags mentioned A 
will 
be conditionally independent from C given D and E. But one can easily check 
that with general type of hypergraph valuation nodes A and C may be rendered 
dependent. 
\begin{th}  \cite{Klopotek:93f}
Hypergraphs considered  by
 Shenoy/Shafer \cite{Shenoy:90} 
may for a given joint belief distribution have simpler structure than
(be properly covered by)
 the closest hypergraph induced by a 
belief network.
\end{th}
%
%
%
%
%
Notably, though the axiomatic system of Shenoy/Shafer refers to hypergraph
factorization of a joint belief distribution, the actual propagation is run 
on a hypertree (or more precisely, on one construction sequence of a 
hypertree, that is on Markov tree) covering that hypergraph \cite{Shenoy:90}. 
\Bem{
Covering a 
hypergraph with a hypertree is a trivial task, yet finding the optimal one 
(with 
as small number of variables in each hyperedge of the hypertree as possible) 
may be very difficult \cite{Shenoy:90}.}
 Let us look closer at the outcome of the process of covering with a reduced
 hypertree factorization, or more precisely, at the relationship of a 
hypertree construction 
sequence and a  belief network constructed out of it in the following way:
If $h_k$ is a twig in the sequence $\{h_1,...,h_k\}$ and $h_{i_k}$ its branch 
with $i_k<k$, then let us span the following directed edges in a belief 
network: First make a complete directed acyclic graph out of nodes 
$h_k-h_{i_k}$. Then add edges $Y_l \rightarrow X_j$ for every $Y_l \in 
h_k \cap h_{i_k}$ and every $X_j \in h_k-h_{i_k}$.
Repeat this for every k=2,..,n. 
\Bem{(Note: no connection is introduced between 
nodes contained  in $h_1$).
}         For k=1 proceed as if $h_1$ were a twig with an
empty set as a  branch for it. 
%
%
 It is easily checked that 
the hypergraph induced by a belief network structure obtained in this way is 
in fact a hypertree (if reduced, then exactly the original reduced 
hypertree). Let us turn now to valuations. 
Let $BEL_i$ be the valuation originally attached to the hyperedge $h_i$. 
then $BEL = BEL_1 \odot ...\odot BEL_n$. 
What conditional belief is to be 
attached to $h_n$ ? First marginalize: $BEL'_n = BEL_1^{\downarrow h_1 \cap 
h_n} \odot \dots \odot BEL_{n-1}^{\downarrow h_{n-1} \cap 
h_n} \odot BEL_n$. 
Now calculate: $BEL"_n={BEL'}_n ^{|h_n \cap h_{i_n}}$, and 
$BEL"'_n={BEL'}_n  ^{\downarrow h_n \cap h_{i_n}}$. 
Let  $BEL_{*k}= BEL_k\oantidot BEL_k ^{\downarrow h_1 \cap 
h_n}$  for k=1,...,$i_n$-1,$i_n$+1,...,(n-1),  
and 
let $BEL_{*i_n}= (BEL_{i_n}\oantidot BEL_{i_n} ^{\downarrow h_1 \cap 
h_n}) \odot BEL"'_n$ .
Obviously, $BEL=BEL_{*1} 
\odot 
\dots \odot BEL_{*(n-1)} \odot BEL"_n$  
Now let us consider a new hypertree only with hyperedges $h_1,\dots 
h_{n-1}$, and 
with valuations equal to those marked with asterisk (*), and repeat the 
process 
until only one hyperedge is left, the now valuation of which is considered 
 as $BEL"_1$. In the process, a new factorization is 
obtained: $BEL=BEL"_1 \odot \dots \odot  BEL"_n$. \\
If now for a hyperedge $h_k$ $card(h_k-h_{i_k})=1$, then we assign $BEL"_k$ 
to 
the node of the belief network corresponding to $h_k-h_{i_k}$. If  for a 
 hyperedge $h_k$ $card(h_k-h_{i_k})>1$, then we split  $BEL"_k$ as follows: 
Let $h_k-h_{i_k}=\{X_{k1},X_{k2},....,X_{km}\}$ and the indices shall 
correspond to the order in the belief network induced by the above 
construction procedure. Then 
$$BEL"_k=BEL ^{\downarrow h_k|h_k \cap h_{i_k}}=  
 \bigodot_{j=1}^{m} BEL ^{\downarrow (h_k \cap h_{i_k}) \cup 
\{X_{k1},...,X_{kj}\} | (h_k \cap h_{i_k}) \cup 
\{X_{k1},...,X_{kj}\}-\{X_{kj}\}}$$
and we assign valuation $BEL ^{\downarrow (h_k \cap h_{i_k}) \cup 
\{X_{k1},...,X_{kj}\} | (h_k \cap h_{i_k}) \cup 
\{X_{k1},...,X_{kj}\}-\{X_{kj}\}}$ to the node corresponding to $X_{kj}$ in 
the network structure. It is easily checked that:
\begin{th} \label{xxxx}    \cite{Klopotek:93f}
(i) The network obtained by the above construction of its structure and 
valuation from hypertree factorization is a belief network.\\
(ii) This belief network represents exactly the joint belief distribution of 
the hypertree\\
(iii) This belief network induces exactly the original reduced hypertree 
structure
\end{th}
The above theorem implies that any hypergraph suitable for 
propagation must have a compatible 
belief network. Hence seeking for belief network decompositions of joint 
belief  distributions  is  sufficient  for  finding  any  suitable 
factorization.

\section{Recovery of Tree-structured Belief Networks}

Let us consider now a special class of hypertrees: connected hypertrees with 
cardinality of each hyperedge equal 2. It is easy to demonstrate that such 
hypertrees correspond exactly to directed trees. Furthermore, valuated 
hypergraphs of this form correspond to belief networks with directed trees as 
underlying dag structures. Hence we can conclude that any factorization in 
form of  connected hypertrees with 
cardinality of each hyperedge equal 2 may be recovered from data by 
algorithms recovering belief trees from data.%
This does not hold e.g. for poly-trees. 

%
Let us assume that there 
exists a measure $\delta(BEL_1,BEL_2)$ equal to zero whenever both belief 
distributions $BEL_1,BEL_2$ are identical and being positive otherwise. 
Furthermore, we assume that $\delta$ grows with stronger deviation of both 
distributions without specifying it further. 
The algorithm of Chow/Liu \cite{Chow:68} for recovery of tree structure of a 
probability distribution is well known and has been deeply investigated, so 
we will omit its  description. 
It requires  a 
distance measure DEP(X,Y) between each two variables X,Y rooted in 
empirical data and 
spans a maximum weight spanning unoriented tree between the 
nodes. Then any orientation of the tree is the underlying dag structure where 
valuations are calculated  as conditional probabilities.  
 To accommodate it for general belief trees one needs a proper measure of 
distance between variables. As claimed  earlier in \cite{Acid:91}, 
this distance measure has to fulfill  the  following 
requirement: 
$ \min(DEP(X,Y),DEP(Y,Z))>DEP(X,Z)$ for any X, Y, 
 Z such that there exists 
a directed path between X and Y, and between Y and Z.
 For probabilistic belief networks one of such functions is known 
to be Kullback-Leibler distance: 
$$    DEP0(X,Y)=\sum_{x,y} P(x,y)\cdot \log 
\frac{P(x,y)} {P(x)*P(y)} 
$$
If we have the measure $\delta$ available, 
we can construct the measure DEP as follows: 
By the ternary joint distribution of the variables  $X_1,X_3$ with background 
 $X_3$ we understand the function:\\
$$BEL ^{\downarrow X_1 \times X_2[X_3]} 
=(BEL ^{\downarrow X_1 \times X_3 | X_3} \odot
 BEL ^{\downarrow X_2 \times X_3 | X_3} \odot BEL ^{\downarrow X_3})
 ^{\downarrow X_1 \times X_2}$$
Then we introduce:\\
$$DEP_{BN}(X_1,X_2)=
 \min(\delta(BEL ^{\downarrow X_1}\odot  BEL ^{\downarrow X_2}, BEL 
^{\downarrow X_1 \times X_2}),$$
$$, \min_{X_3;X_3 \in \V-{X_1,X_2}}
 \quad \delta(BEL ^{\downarrow X_1 \times 
X_2[X_3]}, BEL ^{\downarrow X_1 \times X_2})) $$\\
with {\V}  being the set of all variables.
The following theorem is easy to prove:
\begin{th}  \cite{Klopotek:93f}
$ \min(DEP_{BN}(X,Y),DEP_{BN}(Y,Z))>DEP_{BN}(X,Z)$ for any X, Y, Z such that 
there exists a directed path between X and Y, and between Y and Z.
\end{th}
This suffices to extend the Chow/Liu algorithm to recover general belief tree 
networks from data.

The general algorithm would be of the form:

\begin{itemize}
\item[A)] E be the set of unoriented edges, V be the set of all (at least 3)
variables,
 $V_c$ be the set of connected variables, $V_u$ be the set of not connected
variables. Find the variables X,Y from V maximizing the function 
$DEP_{BN}(X,Y)$ for all pairs (X,Y) of distinct X,Y from V. \\
Initialize: $E=\{(X,Y)\}$, $V_c=\{X,Y\}$, $V_u=V-\{X,Y\}$
\item[B)] repeat\\
Find variables P,Q from V maximizing the function 
$DEP_{BN}(P,R)$ for all pairs (P,R) with $P\in V_c$, $R\in V_u$.\\
Substitute: $E=E \cup \{(P,R)\}$, $V_c=V_c\cup\{R\}$,  $V_u=V_u-\{R\}$.\\
until $V_u$ is an empty set.
\item[C)] Pick one of variables from V and orient all the edges in E away from
this variable.
\end{itemize}
 To demonstrate the validity of this general theorem, its specialization was 
implemented 
for the Dempster-Shafer belief networks. The following $\delta$ function was 
 used: Let    $Bel_1$ be a DS belief function and $Bel_2$ be a DS 
pseudo-belief 
function approximating it. Let 
$$\delta(Bel_2,Bel_1)= \sum_{A; m_1(A)>0} m_1(A) \cdot| \ln 
\frac{Q_1(A)}{Q_2(A)}|$$ 
where the assumption is made that natural logarithm of a non-positive number 
is plus infinity.  $|.|$ is the absolute value operator. The values of 
$\delta$ in variable $Bel_2$ with parameter $Bel_1$ range:
$[0,+\infty)$.
For randomly generated tree-like DS belief distributions, if we were working 
directly with these distributions, as expected, the algorithm yielded perfect 
decomposition into the original tree. For random samples generated from such 
distributions, the structure was recovered properly for reasonable sample 
sizes (200 for up to 8 variables). Recovery of the joint distribution was not 
too perfect, as the space of possible value combination is tremendous and 
probably quite large sample sizes would be necessary. It is worth mentioning, 
that even with some departures  from truly tree structure a distribution 
could be obtained which reasonable approximated the original one.

\section{Recovery of Polytree-structured Belief Networks}

A well known algorithm for recovery of polytree from data for probability 
  distributions is that of Pearl \cite{Pearl:88}, \cite{Rebane:89},  
we refrain from describing it 
here.  To accommodate it for usage with DS belief distributions we had to 
change the dependence criterion of two variables given a third one.

$$Criterion(X_1\rightarrow X_3, X_2 \rightarrow X_3) = 
\delta(BEL ^{\downarrow X_1 \times 
X_2[X_3]}, BEL ^{\downarrow X_1 \times X_2}))
- $$
$$- \alpha \cdot
\delta(BEL ^{\downarrow X_1}\odot  BEL ^{\downarrow X_2}, BEL 
^{\downarrow X_1 \times X_2})
$$
 If the above function  $Criterion$ is negative, we assume
head-to-head meeting 
of edges $X_1,X_3$ and $X_2,X_3$. The rest of the algorithm runs as that of 
Pearl.

The general algorithm would be of the form:

\begin{itemize}
\item[A)] E be the set of unoriented edges, V be the set of all (at least 3)
variables,
 $V_c$ be the set of connected variables, $V_u$ be the set of not connected
variables. Find the variables X,Y from V maximizing the function 
$DEP_{BN}(X,Y)$ for all pairs (X,Y) of distinct X,Y from V. \\
Initialize: $E=\{(X,Y)\}$, $V_c=\{X,Y\}$, $V_u=V-\{X,Y\}$
\item[B)] repeat\\
Find variables P,Q from V maximizing the function 
$DEP_{BN}(P,R)$ for all pairs (P,R) with $P\in V_c$, $R\in V_u$.\\
Substitute: $E=E \cup \{(P,R)\}$, $V_c=V_c\cup\{R\}$,  $V_u=V_u-\{R\}$.\\
until $V_u$ is an empty set.
\item[C)] 
For every pair of edges((X,Z),(Y,Z)) from E sharing an edge end calculate 
$Criterion(X\rightarrow Z, Y\rightarrow Z  )$. (head-to-head) If the result is
negative,
 orient both edges as $(X\rightarrow Z)$, $(Y\rightarrow Z)$, otherwise 
do nothing for them.
\item[D)] Orient remaining unoriented edges as not to cause new head-to-head
meetings.%
\end{itemize}
The steps C) and D) may result in conflicts concerning edge orientations if
the true underlying joint belief distribution was not poly-tree shaped.
Special heuristic procedures need to be applied to resolve them reasonably (In
the actual implementation both steps are in fact intermixed).

For randomly generated polytree-like DS belief distributions, if we were 
working 
directly with these distributions, as expected, the algorithm yielded perfect 
decomposition into the original polytree. For random samples generated from 
such 
distributions, the structure was recovered properly only for very 
large sample 
sizes (5000 for 6 variables), with growing sample sizes leading to 
spurious indications of head-to-head meetings not present in the original 
distribution. Recovery of the joint distribution was also not 
too perfect, due to immense size of  space of possible value combinations.

\section{Recovery of General Type Belief Networks}

Hidden (latent) variables are source of trouble both for 
identification of 
causal relationships (well-known confounding effects) and 
for construction of 
a belief network (ill-recognized direction of causal 
influence may lead to 
assumption of independence of variables not present in the 
real 
distribution). Hence much research has been devoted to 
construction of models 
 with hidden variables. It is a trivial task to construct 
a belief network with hidden 
variables correctly reflecting the measured joint 
distribution. One can 
consider a single hidden variable upon which all the 
measurables depend on. 
But such a model would neither meet the requirements put 
on belief network 
(space saving representation of distribution, efficient 
computation of 
marginals and conditionals) nor those for causal networks 
(prediction 
capability under control of some variables). Therefore, 
criteria like minimal 
latent model (IC algorithm \cite{Pearl:91})  or maximally informative 
partially 
oriented path graph (CI algorithm
and its accelerator FCI algorithm 
 \cite{Spirtes:93}) have been proposed. 
As the IC algorithm 
for learning minimal   latent model  \cite{Pearl:91}  is 
known to be 
wrong \cite{Spirtes:93},
and a failure of FCI has also been reported \cite{Klopotek:93i},
 let us consider the CI algorithm from  
\cite{Spirtes:93}. 

In  \cite{Spirtes:93} the concept of including path graph 
is introduced and 
studied. Given a directed acyclic graph G with the set of 
hidden nodes  $V_h$ 
and visible nodes $V_s$ representing a causal network CN, 
an including path 
between nodes A and  B belonging to $V_s$ is a path in the 
graph G such that  
the only visible nodes (except for A and B) on the path 
are those where edges 
of the path meet head-to-head and there exists a directed  
path  in G from such a node 
to either A or B. An including path graph for G is such a 
graph over $V_s$ in 
which if nodes A and B are connected by an including path 
in G ingoing into A 
and B, then A and B are connected by a bidirectional edge 
$A<->B$. Otherwise 
if they are connected by an including path in G outgoing 
from A and ingoing 
 into B then A and B are connected by an unidirectional 
edge $A->B$. As the set 
$V_h$ is generally unknown, the including path graph (IPG) 
for G is the best we can 
 ever know about G. However, given an empirical 
distribution (a sample), though 
we may be able to detect presence/absence of edges from 
IPG, we may fail to 
decide uniquely orientation of all edges in IPG. 
Therefore, the concept of a 
partial including path graph was considered in 
\cite{Spirtes:93}.  
A partially oriented including path graph contains the 
following types of 
edges unidirectional: $A->B$, bidirectional $A<->B$, 
partially oriented 
$Ao->B$ and non-oriented $Ao-oB$, as well as some local 
constraint information $A*-\underline{*B*}-*C$
 meaning that edges between A and B and 
between B and C cannot meet head to head at B. 
(Subsequently an asterisk (*) 
means any orientation of an edge end: e.g. $A*->B$ means 
either $A->B$ or $Ao->B$ or $A<->B$).
A partial including path graph (PIPG) would be maximally 
informative if all 
definite edge orientations in it (e.g. $A-*B$ or $A<-*B$ 
at A) would be  
shared by all candidate IPG for the given sample and vice 
versa (shared 
definite orientations in candidate IPG also present in 
maximally informative 
PIPG), the same should hold for local constraints. 
Recovery of the maximally informative PIPG is considered 
in \cite{Spirtes:93} 
as too ambitious and a less ambitious algorithm CI has 
been developed therein 
 producing a PIPG where only a subset of edge end 
orientations of the maximally 
 informative PIPG are recovered. Authors of CI claim such 
an output to be 
still useful when considering direct and indirect causal 
influence among 
visible variables as well as some prediction tasks. 

However, CI algorithm is known to be of high computational complexity even for
probabilistic variables. Therefore, we developed a modified version of it
\cite{Klopotek:94}, \cite{Klopotek:93h} to reduce its complexity and to
provide a bridge towards application for DS empirical distributions. \\

%

 We cite below some useful definitions from 
\cite{Spirtes:93} and then present our Fr(k)CI algorithm.

 In a partially oriented including path graph $\pi$:
\begin{itemize}
\item[(i)] A is a parent of B if and only if edge $A->B$ 
is in $\pi$.
\item[(ii)] B is a collider along the path $<A,B,C>$ if 
and only if $A*->B<-*C$ in $\pi$.
\item[(iii)] An edge between B and A is into A iff $A<-*B$ 
is in $\pi$
\item[(iv)] An edge between B and A is out of A iff $A->B$ 
is in $\pi$.
\item[(v)] In a partially oriented including path graph 
$\pi$, U is a definite 
 discriminating path for B if and only if U is an 
undirected path between X and 
Y containing B, $B \neq X, B \neq Y$, every vertex on U 
except for B and the 
endpoints is a collider or a definite non-collider on U 
and:\\
(a) if V and V" are adjacent on U, and V" is between V and 
B on U, then $V*->V"$ on U,\\
(b) if V is between X and B on U and V is a collider on U, 
then $V->Y$ in $\pi$, else $V<-*Y$ on $\pi$\\
(c) if V is between Y and B on U and V is a collider on U, 
then $V->X$ in $\pi$, else $V<-*X$ on $\pi$\\
(d) X and Y are not adjacent in $\pi$.\\
(e) Directed path U: from X to Y: if V is adjacent to X on 
U then $X->V$ in 
$\pi$, if $V$ is adjacent to Y on V, then $V->Y$, if V and 
V" are adjacent on U 
and V is between X and V" on U, then $V->V"$ in $\pi$.
\end{itemize}%

%
%
Let us introduce some notions specific for Fr(k)CI:
\begin{itemize}
\item[(i)] A is r(k)-separated from B given set S 
($card(S)\leq k$) iff A and 
B are conditionally independent given S 
\Bem{- conditional independence means  
$\chi 
 ^2$-test does not deny the thesis of independence of 
variables A and B given S. .}
\item[(ii)] In a partially oriented including path graph 
$\pi$, 
 a node A is called {\em legally removable} iff there 
exists no local constraint 
 information $B*-\underline{*A*}-*C$ for any nodes B and C 
and there exists no
edge of the form $A*->B$ for any node B. 
\end{itemize}
{\noindent \bf The Fast Restricted-to-k-Variables Causal 
Inference Algorithm (Fr(k)CI):}\\
Input: Empirical joint probability distribution\\
Output: Belief network.
\begin{description}
\item[A)] Form the complete undirected graph Q on the 
vertex set V.
\item[B')] 
for j=0 step 1 to k\\ 
do
if A and B are r(k)-separated given any subset S  of 
neighbours of A or of
B, card(S)=j, remove the edge between 
A and B, and record S in Sepset(A,B) and Sepset(B,A). 
\item[B'')] if A and B are r(k)-separated given any subset 
S of V ($card(S)>0$), remove the edge between 
A and B, and record S in Sepset(A,B) and Sepset(B,A). 
\item[C)] Let F be the graph resulting from step B). 
Orient each edge as 
$o-o$ (unoriented at both ends).  For each 
triple of vertices A,B,C such that the pair A,B and the 
pair B,C are each 
adjacent in F, but the pair A,C are not adjacent in F, 
orient \Bem{$(C)$} A*-*B*-*C as 
$A*->B<-*C$ if and only if B is not in Sepset(A,C), and 
orient A*-*B*-*C 
as $A*-\underline{*B*}-*C$ if and only if B is in 
Sepset(A,C).
\item[D)] Repeat
\begin{description}
\item[(D1) if] there is a directed path from A to B, and 
an edge  A*-*B, orient \Bem{$(D_p)$} A*-*B 
as $A*->B$,
\item[(D2) else if]  B is a collider along $<A,B,C>$ in 
$\pi$, B is adjacent 
to D, A and C are not adjacent, and there exists 
 local constraint  $A*-\underline{*D*}-*C$, then orient 
\Bem{$(D_s)$} $B*-*D$ as $B<-*D$ ,
\item[(D4) else if] $P*-\underline{>M*}-*R$ then orient 
\Bem{$(D_c)$} as $P*->M->R$.
\item[(D3) else if] U is a definite discriminating path 
between A and B for M in $\pi$ and 
P and R are adjacent to M on U, and P-M-R is a triangle, 
then\\
if M is in Sepset(A,B) then M is marked as non-collider on 
subpath $P*-\underline{*M*}-R$\\
else $P*-*AM*-*R$ is oriented \Bem{$(D_d)$} as 
$P*->M<-*R$,
\item[until] no more edges can be oriented.
\item[E)] Orient every edge $Ao->B$ as $A->B$.
\item[F)] 
\Bem{
 Orient all the edges of type $Ao-oB$ either as $A<-B$ or 
$A->B$ so as not to 
violate  $P*-\underline{*M*}-*R$ constraints as follows: 
}
Copy the  partially oriented including path graph $\pi$ 
onto $\pi'$. \\
Repeat: \\
 In $\pi'$ identify a legally removable node A. Remove it 
from  $\pi'$ 
together with every edge $A*-*B$ and every constraint 
 with A involved in it. Whenever an edge $Ao-oB$ is 
removed from $\pi'$, orient 
edge $Ao-oB$ in $\pi$ as $A<-B$. \\
Until no more node is left in $\pi'$.
\item[G)]  Remove     every bidirectional edge $A<->B$ and 
insert instead
parentless hidden variable $H_{AB}$ adding edges 
$A<-H_{AB}->B$ 

 \end{description}
\item[End of Fr(k)CI]
\end{description}
The algorithm Ci  has been  first of all moved to DST grounds by using DST
independence tests instead of probabilistic ones.

Steps E) and F) constitute an extension of \Bem{(are not 
present in)} the original CI algorithm of 
\cite{Spirtes:93}, bridging the gap between partial 
including path graph and the belief network (see also \cite{Klopotek:93g}). 

Conditional belief functions, also in presence of hidden variables are
calculated according to \cite{Klopotek:93h}.

Step B)  was modified by substituting the term 
"d-separation" with 
"r(k)-separation" \cite{Klopotek:94}. This means that not all possible 
subsets S of the set of 
all nodes V (with card(S) up to card(V)-2) are tested on 
rendering nodes A and 
B independent, but only those with cardinality 
0,1,2,...,k. If one takes into 
account that higher order conditional independencies 
require larger amounts of 
data to remain stable, superior stability of this step in 
Fr(k)CI becomes 
obvious. Furthermore, this step was subdivided into two 
substeps, B') and
B"). The first substep corresponds to technique used by 
FCI - restriting
candidate sets of potential d-separators to the so far 
established
neighbourhood. This substep is followed by the full search 
over all nodes of
V - but only for edges left by B' - this is in contrast to 
FCI which omits
step B) of the original CI, and thus runs into the 
troubles described in \cite{Klopotek:93i}. 
 
Step D2) has been modified in that the term "not 
d-connected" of CI was substituted 
by reference to local constraints. In this way results of 
step B) are 
 exploited more thoroughly and in step D) no more 
reference is made to original 
body of data (which clearly accelerates the algorithm). 
This modification is 
legitimate since all the other cases covered by the 
concept of  "not 
d-connected" of CI would have resulted in orientation of 
$D*->B$ already in 
step C). Hence the generality of step D2) of the original 
CI algorithm is not needed here. 

Steps D3) and D4) were interchanged as the step D3) of CI 
is quite time
consuming and should be postponed until no alternative 
substep can do anything.\\

\section{Discussion}

The particular feature of the presented algorithms for identification of
belief structure in DST is the close relationship between target knowledge
representation scheme, the reasoning scheme and the data viewing scheme. This
unity enabled to adopt some probabilistic belief network recovery algorithms
for purposes of DS belief network recovery from data. 
If, for example, no frequentist interpretation for DST existed (as required by
Smets \cite{Smets:92}), then the last algorithm (Fr(k)CI) would be pointless,
as it relies on statistical tests over a sample. Furthermore, if the reasoning
scheme would not be that of Shenoy-Shafer \cite{Shenoy:90}, then clearly
tree-generation and polytree generation algorithms would make little sense as
the knowledge representation scheme might require more general 
belief functions
than that
utilized by Shenoy and Shafer. It may prove also pointless to represent a
joint belief distribution as a belief network as 
local influences among variables may lose any meaning
(being replaced by some global fixpoint state of all variables). 
 One should also notice that the possibility of
utilization of statistical independence test for simplification of a joint
belief distribution (decomposition of it) is bound to separate measurability
of attributes. But this separate measurability  is obvious for Shenoy-Shafer
uncertainty propagation scheme, but does not need to be so for other.

On the other hand, the notion of statistical independence is strongly related
with the interpretational attitude of the researcher. E.g. if we consider DST
as calculus over lower and upper probability bounds, as done e.g. in
\cite{Halpern:92}, then not only the reasoning scheme would have to be changed
(is indicated in \cite{Halpern:92}) but also the understanding of independence
of variables: instead of missing mutual influence the possibility of missing
mutual influence would have to be considered. The application of Chow/Liu
algorithm \cite{Chow:68} would then be connected with notion of lower
and upper entropy etc. \\
Some researchers do not even bother if their notion of independence has at all
a empirical sense. E.g. Hummel and  Landy \cite{Hummel:88}   talk about
"independent opinions of experts" within their probabilistic interpretation.
But how such an independence has to be understood ?  Opinions  of 
experts on a
subject cannot be independent as they have a common cause  -  the 
subject for
which these opinions are issued. Should opinions of experts be really
independent, then at least one of the opinions would have to be unrelated to
the subject of expertise, hence devoid of any useful content. Another
strange approach is exhibited in a recent paper by Zhu and
Lee \cite{Zhu:93}
where it is a priori assumed that premise and conclusion of an expert rule are
statistically independent. Under these circumstances the value of a
reasoning rule and hence of the whole reasoning system is questionable (we
can infer a posteriori beliefs without knowledge of observables). 

One of the few interpretations of belief functions possessing intrinsic
physical relevance is the rough set based interpretation presented by Skowron
\cite{Skowron:93} and Grzyma{\l}a-Busse \cite{Grzymala:91}. This rough set
approach explains a possible physical source  of  Dempster-Shafer 
uncertainty
(incomplete observable set for decision table). However, it 
couldn't be used in
combination with our algorithms as it does not fit the separate measurability
requirement (enforcing separate measurability would cause loss of 
information
otherwise present in decision table). Rough set approach allows also for more
precise representation of conditional relationships between decision variables
than that actually imposed by Shafer's conditioning. Therefore Shenoy-Shafer
uncertainty propagation scheme is also not suitable for application within the
rough set framework as it would deteriorate decision table capabilities.

The algorithms developed constitute in some sense extensions of known
algorithms from the bayesian belief network literature. However, these
extensions were  not  straightforward  ones.  First  of  all  the 
empirical meaning
of independence and conditional independence from data for DS theory had to be
established. This required imposition of a compatible probabilistic
interpretation of DST. However, as just stated,  no such completely useful
interpretation was available from literature. Hence one had to be developed.
As the tree-recovery algorithm is concerned, the general famework of Chow/Liu
algorithm \cite{Chow:68} could be adopted with the exception of distance
measure. This measure had to be invented completely from scratch as the
intuition behind Chow/Liu original measure is that of probabilistic
composition which has properties contrary to DS composition (compare the
effects of decompositions). Similarly conditional distance had to be designed
anew for polytree-recovery algorithm of Pearl \cite{Rebane:89}. \\
In case of general network with variable hiding the adaptation of CI algorithm
Spirtes et al. \cite{Spirtes:93} involved more complex work. As 
stated already,
the notion of conditional independence from data for DST had to be invented.
Furthermore, a result in form of a belief network instead of CI's partial
including path graph was required \cite{Klopotek:93g} and later adopted for
DST \cite{Klopotek:93h}. But the time complexity of CI was already high for
probabilistic networks and explodes for DS networks. Therefore some
simplifications and algorithmic improvements had to be carried out (compare
\cite{Klopotek:94}). Last not least, some heuristics for calculation of
marginal and conditional  distributions  from  data  for  and  in 
presence of hidden
variables had to be elaborated (This task was not reported here).


\section{Conclusions}

Within this paper belief network discovery algorithms for three different
classes of Dempster-Shafer
belief networks (tree-structured, poly-tree-structured, general ones with
variable hiding) have been presented. Close relationship between utility of
these algorithms and the usage of a particular uncertainty propagation scheme
(Shenoy-Shafer local computation method \cite{Shenoy:90}) has been
demonstrated. Also a new frequentist interpretation of DST has been described
and shown to be a prerequisite for application of  developed algorithms.
Though in basic ideas these algorithms resemble ones known from bayesian
network literature, considerable effort was required for clearing various
details,  like   measures   of   distances   between   variables, 
instantiating of causal
networks to belief networks etc.\\

It is hoped that this research may contribute to adaptation of further
bayesian network recovery algorithms for DS belief networks and/or outline
procedures for development of complex probabilistic models for other known
uncertainty propagation schemes of DST.

\newcommand{\eng}[1]{}    
\newcommand{\pol}[1]{}    
\newcommand{\deu}[1]{}    
\newcommand{\A}[2]{#2 #1} 

\renewcommand{\eng}[1]{#1}    
\newcommand{\IN}{\pol{[w:]}\deu{[in:]}\eng{[in:]}} 
\newcommand{\LitStelle}[2]{\bibitem{#1} }   

\newcommand{\ReadingsIn}{G. Shafer, J. Pearl eds: {\it Readings in Uncertain 
Reasoning}, (ISBN 1-55860-125-2, 
Morgan Kaufmann Publishers Inc., San Mateo, California, 1990)}

\end{document}